\documentclass[conference]{IEEEtran}
\IEEEoverridecommandlockouts
\usepackage[numbers, sort&compress]{natbib}
\usepackage{amsmath,amssymb,amsfonts}
\usepackage{graphicx}
\usepackage{booktabs}
\usepackage[dvipsnames]{xcolor}
\newenvironment{varalgorithm}[1]
  {\algorithm\renewcommand{\thealgorithm}{#1}}
  {\endalgorithm}
  
\usepackage{tensor}
\usepackage{algorithm}
\usepackage{algpseudocode}
\usepackage{adjustbox}
\usepackage{array}
\usepackage{pifont}
\usepackage{multirow}
\usepackage[T1]{fontenc}
\usepackage{indentfirst}
\usepackage{ragged2e}

\newcommand*\rot{\rotatebox{45}}
\newcommand*\OK{\ding{51}}
\newcommand*\notOK{\ding{55}}

\newcolumntype{P}[1]{>{\Centering\hspace{0pt}}p{#1}}
\newcolumntype{L}[1]{>{\RaggedLeft\hspace{0pt}}l{#1}}

\usepackage{epsfig}
\usepackage{latexsym}
\usepackage{amsfonts}
\usepackage{here}
\usepackage{rawfonts}
\usepackage[latin1]{inputenc}
\usepackage{calc}

\usepackage{url}
\usepackage{enumerate}
\usepackage{color}

\usepackage{amssymb}
\usepackage{upref}
\usepackage{epic,eepic}
\usepackage{times}
\usepackage{dsfont}
\usepackage{comment}
\usepackage{mathrsfs}
\usepackage{verbatim}
\usepackage{float}
\usepackage{chngcntr}
\usepackage{bbm}

\usepackage{hyperref}
\usepackage{subcaption}
\usepackage[english]{babel}
\usepackage{blindtext}
\allowdisplaybreaks

\DeclareMathOperator*{\argmax}{argmax}

\newtheorem{proposition}{\bf Proposition}
\let\oldproposition\proposition
\renewcommand{\proposition}{\oldproposition\normalfont}

\newtheorem{definition}{\bf Definition}
\let\olddefinition\definition
\renewcommand{\definition}{\olddefinition\normalfont}

\newlength{\totlinewidth}
\newcounter{substep}

  {\end{list}}

\newlength{\aligntop}
\newlength{\alignbot}
 \makeatletter
\renewenvironment{align}{%
  \vspace{\aligntop}
  \start@align\@ne\st@rredfalse\m@ne
}{%
  \math@cr \black@\totwidth@
  \egroup
  \ifingather@
    \restorealignstate@
    \egroup
    \nonumber
    \ifnum0=`{\fi\iffalse}\fi
  \else
    $$%
  \fi
  \ignorespacesafterend%
  \vspace{\alignbot}\par\noindent
} \makeatother

\begin{document}

\title{Triplet Loss-less Center Loss Sampling Strategies in Facial Expression Recognition Scenarios\\
\thanks{This material is based upon work supported by the Air Force Office of Scientific Research under award number FA9550-20-1-0090 and the National Science Foundation under Grant Numbers CNS-2232048, and CNS-2204445.}
\vspace{-0.2cm}}


\author{
	\IEEEauthorblockN{
	Hossein Rajoli\IEEEauthorrefmark{1}, Fatemeh Lotfi\IEEEauthorrefmark{1}, 
 Adham Atyabi\IEEEauthorrefmark{2},
 Fatemeh Afghah\IEEEauthorrefmark{1}}

	\IEEEauthorblockA{\IEEEauthorrefmark{1}Holcombe Department of Electrical and Computer Engineering, Clemson University, Clemson, SC, USA \\
Emails: \{hrajoli, flotfi, fafghah\}@clemson.edu,}
\IEEEauthorblockA{\IEEEauthorrefmark{2}Department of Computer Science, University of Colorado Colorado Springs, Colorado Springs, USA \\
	 Email: aatyabi@uccs.edu}}

\maketitle \vspace{-0.1cm}

\begin{abstract}
Facial expressions convey massive information and play a crucial role in emotional expression. Deep neural network (DNN) accompanied by deep metric learning (DML) techniques boost the discriminative ability of the model in facial expression recognition (FER) applications. DNN, equipped with only classification loss functions such as Cross-Entropy cannot compact intra-class feature variation or separate inter-class feature distance as well as when it gets fortified by a DML supporting loss item. The triplet center loss (TCL) function is applied on all dimensions of the sample's embedding in the embedding space. In our work, we developed three strategies: fully-synthesized, semi-synthesized, and prediction-based negative sample selection strategies. To achieve better results, we introduce a selective attention module that provides a combination of pixel-wise and element-wise attention coefficients using high-semantic deep features of input samples. We evaluated the proposed method on the RAF-DB, a highly imbalanced dataset. The experimental results reveal significant improvements in comparison to the baseline for all three negative sample selection strategies.
\end{abstract}


\section{Introduction}
Facial expression recognition (FER) has always played an essential role in the development and flourishing of numerous daily-used applications. Autonomous cars, human-computer interaction (HCI), quality of experiment measurement, forensic section, healthcare, and treatment, detecting deception, etc. are fields in which automatic FER is widely used as monitoring or assessment tools \cite{revina2021survey}. Increasing computational capacity, emerging sophisticated deep neural networks (DNN), and providing large-scale datasets have made great progress in this hot spot field of computer vision. Expressed emotions by face depend on the main three categories of signals proposed by the face, namely, static, slow, and fast signals. All discriminative features fall into these three signals, for instance, skin color, and facial bone shape of individuals belong to the static group, while muscle tone and skin texture, and eyebrow rapid rising respectively belong to slow and fast signals. Therefore, in the same category of facial emotional expression, there are high diversity of samples that vary due to age, level of intelligence, race, gender and so on \cite{revina2021survey, mehrabi2021age}.

In general, there are three main steps in a standard FER task with the DNN; extracting spatial deep features of the input sample, mapping highly semantic feature space to embedding space, and then estimating the probability distribution over all categories using softmax activation. Meanwhile, deep metric learning (DML) can play a crucial role. By imposing embedding space toward the point that intra-class samples are grouped tightly and inter-class ones placed as far as possible, DML can improve our supervised classification task \cite{kim2021embedding}. One of the effective DML-based loss functions is loss-less triplet loss \cite{kertesz2021different} which has been designed to take the information of all samples in backpropagation. 

In an ordinary triplet loss, if the distance between the embedding space's sample representation and its corresponding center, which represents a positive sample, is less than its distance to the negative ones with the amount of defined margin, then that sample would not contribute to backpropagation. To boost the performance, a particular form of TC3L, namely, \emph{adaptive margin T3L (AMTC3L)}, is introduced. Compared with ordinary TC3L, which considers a constant margin for all inputs, AMTC3L applies adaptive margins for each sample. The proposed loss function focuses on highlighting the samples with features that improve the discriminative ability of the network and let them contribute more to the backpropagation process.


Two different attention modules are introduced, pixel-wise and element-wise inclusion/exclusion modules. The core idea is, only distinguishing features from discriminative spots of the face get filtered out for recognizing the expressed emotion. In this way, the model learns how to discriminate noisy features and areas based on the deep features that have been extracted.

The main contributions of this work can be summarized as follows;
\begin{itemize}
  \item We proposed a new form of triplet loss (TL), namely, AMTC3L, where the network learns the appropriate margin that well suits the network and lets it learn from some samples more than others by assigning appropriate margin to each sample attentively. 
  
  \item To emphasize more on discriminative features we proposed an attention network, yields exclusion/inclusion weights from the deep features. 
  \item To apply AMTC3L to the FER scenario, we proposed three different strategies offering the most negative sample. They are full-synthesized, semi-synthesized, and prediction-based negative sample selection strategies.
  \item The source code will publicly be available at the following github repository: \href{https://github.com/HRajoliN/Triplet-Loss-less-Center-Loss-Sampling-Strategies-in-FER}{AMTC3L sampling strategies.}
\end{itemize}




\section{Related work}
This paper mainly focuses on introducing three strategies to customize and leverage triplet loss in FER scenarios fortified by a DML approach. Hence, we consider conducted works from two perspectives of DML usage in FER and triplet loss sampling strategies in classification scenarios.

\subsection{Triplet Loss sampling strategies in classification}
Triplet loss enhances both inter-class compactness and intra-class separability.
The authors in \cite{2021triplet} developed a class-paired margin identifier mechanism in which the association with an outlier detector improves the performance of FER application. They proposed an outlier detector preventing occluded and confusing samples from having contributions in selected negative samples. To classify multi-labeled samples, \cite{2022informative} proposed a two-step sample selection mechanism. In the first step, a subgroup of the most informative diverse samples have been selected from the mini-batches as anchors and then in the second step, the positive and negative samples would be chosen considering the relevancy, hardness, and diversity of the images during the selection. To address the 3D retrieval challenge, \cite{2018triplet} leveraged triplet center loss (TCL) and introduced a class center as the anchor sample, and adopted the nearest negative sampling as the negative sample selection strategy. Since the triplet loss increases the chance of contribution in back-propagation for minority classes in an imbalance dataset as the negative sample, \citet{2020class} takes this advantage to classify the imbalance dataset with triplet and triplet center loss functions. The authors adopt all class centers as the negative sample, not only the nearest ones in their classification scenario. \citet{2021hyperspectral} conducted research to address the challenges of classifying hyperspectral images that are prone to inter-class inseparability due to occupying a wide range in feature space. To leverage the triplet loss function, in each mini-batch they randomly select a subset of classes and for each selected class they randomly take a subset of their corresponding samples within the mini-batch. Selecting a set of the hardest positive and negative samples for each input prevents outliers to affect the training procedure. Triplet loss has been also used to address knowledge distillation challenges. \citet{2021teacher} distilled and transferred knowledge of a trained network on the full-face dataset to a network that has been trained on the occluded-face dataset. In the proposed scenario, the occluded sample in embedding space and the negative samples come from the student network while the full-face networks give the positive samples. In both the positive and negative sampling process, the distance between the anchor and one-tenth of the randomly selected available samples is considered. Then, the farthest and the nearest samples are chosen as the positive and negative samples. \citet{2022discriminative} utilized TCL with a fixed margin in a FER scenario.
In this work, the input sample's center is considered as the positive sample, and the nearest class center (other than the sample's class center) as the negative one. Chi-squared distance has been applied to the triplet loss function in a facial emotion recognition scenario by \cite{2022histnet}. The authors considered an identity annotation (if available) in negative sample selection to reduce the computation cost and the effect of identity features in the training process.

\subsection{DML in FER scenarios}
Deep metric learning using a neural network, is an approach for projecting input samples into a manifold space where a distance metric such as euclidean, cosine, or chi-squared can measure the distance between two points \cite{2018deep}. For facial emotion recognition, \citet{2022ad} proposed an adaptive correlation loss that guides the network within the training process to the point of being more discriminative. They define an adaptive loss function based on the mean of $k$ embedding vectors that the network produces in the embedding space for every single class. The loss function penalizes less correlated intra-class mean vectors and correlated inter-class ones. \citet{farzaneh2021facial} takes advantage of center loss with an attentive network to increase the intra-class compactness. \citet{2021identity} developed a Mahalanobis metric distance and implemented it by a fully connected network, applied to the higher semantic features. To reach the level of pose-aware and identity-invariant, authors of \cite{2021dynamic} proposed a five-tuple DML consisting of two sets of triplets including, pos triplet and identity triplet with a common anchor, then trained the network using this DML approach. \citet{2017adaptive} proposed $(N+M)$-tuplet clusters' loss that selects $N$ different expressions from the same identity as the negative samples and searches for $M$ positive samples on the fly then optimizes the network jointly based on this loss plus softmax  to alleviate attribute variations coming from different identities.

\begin{figure*}[htbp]
\centerline{\includegraphics[width=2\columnwidth]{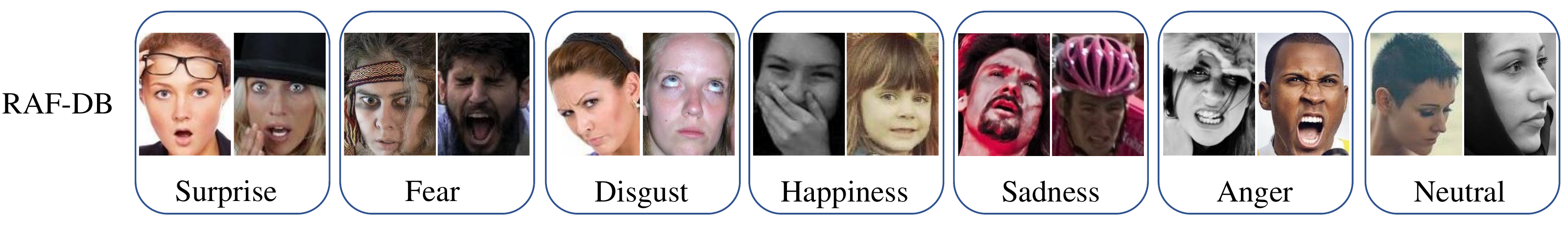}}
\caption{A sample of RAF-DB facial expressions dataset.}\vspace{-0.2cm}
\label{fig2}
\end{figure*}

\section{Proposed model}
In this section, we elaborate on the proposed approach and explain all modules, namely, Resnet-50 as the backbone, the classifier, the multi-task loss function, purifying module that eliminates outlier features from the embedding vector while providing the adaptive margin for input samples. In our approach, using weights generated by purifying network the margin would be calculated as, $0 \leq \{\alpha_i = \sum_{j=1}^{C_d} w_j \} \leq C_d$.
Given a mini batch of $m$ samples- $D_m = \{ (x^{(i)} , y^{(i)}) \mid i = 1 , 2 , \ldots , m \}$, where $x^{(i)} \in \mathbb{X}$ is the training sample from the training set and $y^{(i)} \in \{1 , 2 , \ldots , K\} $ is its corresponding categorical label. The backbone network, $\mathcal{N_{BB}}$, maps mini batch input samples to their feature space representation, $x^{\ast} \in \mathbb{R}^{C_{f} \times H_{f} \times W_{f}}$, where $C_{f}$, $H_{f}$, and $W_{f}$ respectively denote the number of channels, rows, and width of the tensor in the feature space.
By leveraging two depth-wise convolutional layers, the classifier network, $\mathcal{N_C}$, takes the feature space representation of the sample and extracts deep high semantic context, $x^d \in \mathbb{R}^{C_{d} \times H_{f} \times W_{f}}$. Then, using a global average pooling layer, $\mathcal{P}$, the tensor gets converted to the embedding $e \in \mathbb{R}^{C_{d} \times 1 \times 1}$. Finally, another depth-wise convolutional layer maps the embedding to row score of $x^z \in \mathbb{R}^{K \times 1 \times 1}$. Then, a probability distribution over $K$ channels, $Pr(y = j \mid x^{(i)})$, is taken using Softmax over channels. In the end, the discrepancy between prediction, $\hat{y}$ and true labels, $y^{(i)}$, would be calculated using cross-entropy (CE) loss function, $\mathcal{L}_{CE}$, as follows:\vspace{-0.2cm}

\begin{equation}
\label{CE_loss}
\mathcal{L}_{CE} = -\frac{1}{m} \sum_{i = 1}^{m} \sum_{j = 1}^{K} \log y^{(i)} Pr(y^{(i)} = j \mid x^{(i)}).
\end{equation}

\begin{figure*}[htb]
\centerline{\includegraphics[width=1.8\columnwidth]{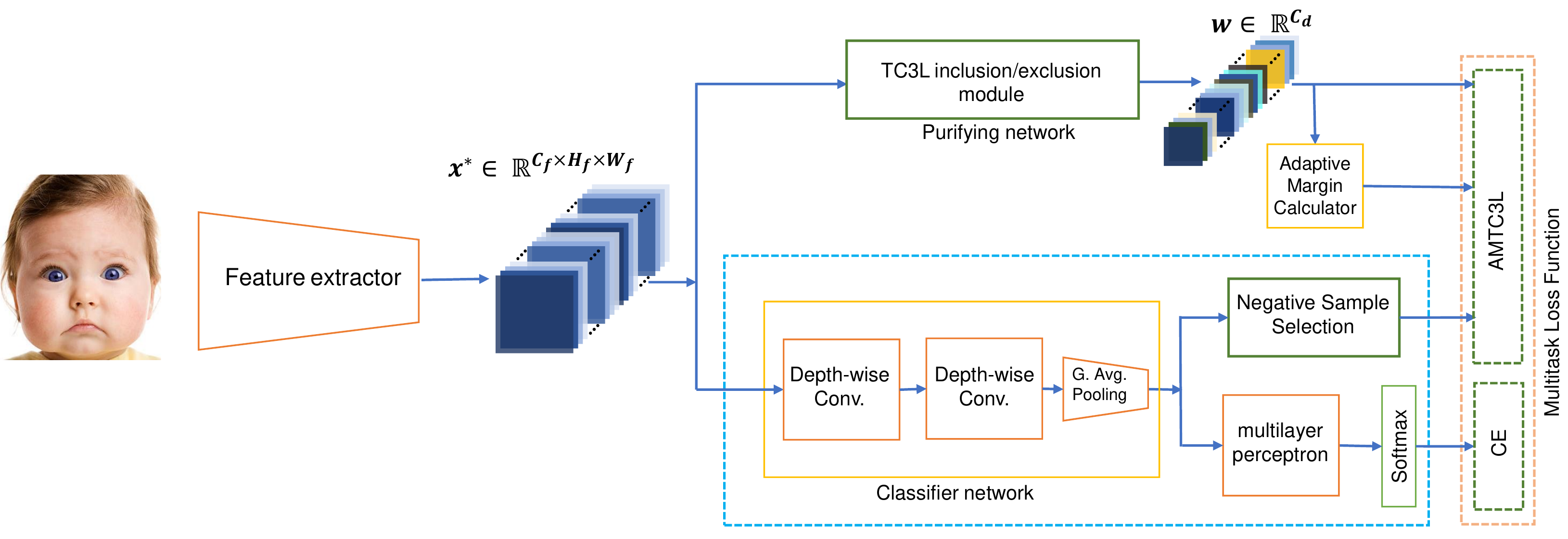}}
\caption{Detailed structure of the model.}\vspace{-0.2cm}
\label{Emphasizer}
\end{figure*}

\subsection{Multi-task Loss}
To calculate discrepancy, for any training sample, the following loss function would be calculated:
\begin{equation}
\label{multi_task_loss}
\mathcal{L} = \mathcal{L}_{CE} + \lambda \, \mathcal{L}_{AMTC3L}, 
\end{equation}
where \eqref{CE_loss} calculates $\mathcal{L}_{CE}$, and $\mathcal{L}_{AMTC3L}$ measures Adaptive margin triplet center loss.


\subsubsection{Adaptive margin triplet loss-less center Loss (AMTC3L)}
The functionality of AMTC3L relies on two sub-modules, the margin estimator, and the negative class-center selector. As explained in \cite{schroff2015facenet} and demonstrated in \eqref{vanilla_TL}, in general, the marginal triplet loss has three main terms that should be chosen carefully, the positive sample, the negative sample, and the margin. To improve intra-class compactness and inter-class separability, the sample's class center in embedding space is considered as the positive sample and the negative one would be selected from the remaining $k-1$ class centers.
\begin{align}
\label{vanilla_TL}
\mathcal{L}_{TL} = \, \max \Big(0 \, ,& \, \frac{1}{2m} \sum_{i=0}^{m} \sum_{j=1}^{c_d} {\| {(e^{(i)}_{j})} - (c^{p}_{y^{(i)}{j}}) \| _{2}^{2}} \nonumber\\ 
& -{\| ({e^{(i)}_{j}}) - (c^{n}_{y^{(i)}{j}}) \| _{2}^{2}} + \alpha_i \Big),
\end{align}
where, $\sigma$ represents the sigmoid activation, $e^{(i)}_{j} \in e^{(i)} = [e^{(i)}_1 , e^{(i)}_2 , \ldots , e^{(i)}_d]^T$ represents $j'th$ element of the $i'th$ sample's embedding from the input mini batch. Corresponding class center, $c^{p}_{y^{(i)}{j}} \in c^{p}_{y^{(i)}} = [c^{p}_{y^{(i)}1} , c^{p}_{y^{(i)}2} , \ldots , c^{p}_{y^{(i)}d}]$ defines the most positive embedding, while $c^{n}_{y^{(i)}{j}} = [c^{n}_{y^{(i)}1} , c^{n}_{y^{(i)}2} , \ldots , c^{n}_{y^{(i)}d}]$ represents the class center of the most negative one. Note that $\alpha_i$ is the margin of sample $i$th of minibatch which comes from the adaptive margin generator.
\begin{varalgorithm}{MS-NSS:}
\footnotesize
\caption{Mathematically Synthesized Negative-Sample Selection}
\label{negative_sample_selection_first}
\textbf{Inputs:}\,mini batch's embedding tensors and corresponding labels,\\
$D_m = \{(e^{(i)},y^{(i)}) \mid i=0,1,...,m \, ; \, e^{(i)} \in \mathbb{R}^{C_{d} \times 1 \times 1} \}$;\\
embedding tensors' centers,\\ 
$c = \{ c_k \mid k = 1 , 2 , \ldots , K \, ; \, c_{(k)} \in \mathbb{R}^{1 \times C_{d}} \}\}$. \\
\textbf{Output:}\,\, $c^n = \{ c^n_i \mid i = 1 , 2 , \ldots , m \}$.
\begin{algorithmic}[1]
\While {not converged}

\State compute inter cluster euclidean distance:  

$\eta_{kj} = \{{\| \sigma(({e^{(i)}_{j}}) - \sigma((c_{c_{k}{j}}) \| _{2}^{2}} \mid i = 1,...,m \,$

\hspace{7mm} $ ; j = 1 , \ldots , C_d$

\hspace{7mm} $; \, \forall \, c_k \mid k \neq y^{(i)}) \, , \, k = 1 , 2 , \ldots , K \}$.
\State  $c^{n}_j = \{ \underset{k \in \{1,2,\ldots,C_d\}}{\text{minimize}} \,\, (\eta_{kj}) \mid j = 1,2,\ldots,C_d\}$
\State compute the triplet center loss by \eqref{AMTCL}.
\EndWhile
\end{algorithmic} 
\end{varalgorithm}\setlength{\textfloatsep}{.2\baselineskip}


\begin{varalgorithm}{NS-NSS:}
\footnotesize
\caption{Network's Semantic Negative Sample Selection}
\label{negative_sample_selection_second}
\textbf{Inputs:}\,mini batch's embedding tensors and corresponding labels,\\
$D_m = \{(e^{(i)},y^{(i)}) \mid i=0,1,...,m \, ; \, e^{(i)} \in \mathbb{R}^{C_{d} \times 1 \times 1} \}$;\\
mini batch's prediction, $\hat{y}^{(i)}$;\\
embedding tensors' centers,\\ 
$c = \{ c_k \mid k = 1 , 2 , \ldots , K \, ; \, c_{(k)} \in \mathbb{R}^{1 \times C_{d}} \}\}$. \\
\textbf{Output:}\,\, $c^n = \{ c^n_i \mid i = 1 , 2 , \ldots , m \}$.
\begin{algorithmic}[1]
\While {not converged}

\If{$\hat{y}^{(i)} \neq y^{(i)}$}:

\State ${c^{n}} = \{ c_{\hat {y}^{(i)}} \mid i \in \{i \mid \hat{y}^{(i)} \neq y^{(i)} \} \}$

\Else \,\,{$\hat{y}^{(i)} = y^{(i)}$}:
\State $\gamma ^{(i)} = \argmax \{ \upsilon(\hat{y}^{(i)} = k) \mid k \neq y^{(i)} \}$

\hspace{2mm} here $\upsilon()$ represents frequency function.

\State ${c^{n}} = \{ c_{\gamma ^{(i)}} \mid i \in \{i \mid \hat{y}^{(i)} = y^{(i)} \} \}$

\EndIf

\State compute the triplet center loss by \eqref{AMTCL}.
\EndWhile
\end{algorithmic} 
\end{varalgorithm}\setlength{\textfloatsep}{.2\baselineskip}
\begin{varalgorithm}{MM-NSS:}
\footnotesize
\caption{Mixed-Method Negative Sample Selection}
\label{negative_sample_selection_third}
\textbf{Inputs:}\,mini batch's embedding tensors and corresponding labels,\\
$D_m = \{(e^{(i)},y^{(i)}) \mid i=0,1,...,m \, ; \, e^{(i)} \in \mathbb{R}^{C_{d} \times 1 \times 1} \}$,\\
embedding tensors' centers,\\ 
$c = \{ c_k \mid k = 1 , 2 , \ldots , K \, ; \, c_{(k)} \in \mathbb{R}^{1 \times C_{d}} \}\}$. \\
\textbf{Output:}\,\, $c^n = \{ c^n_i \mid i = 1 , 2 , \ldots , m \}$.
\begin{algorithmic}[1]
\While {not converged}

\If{$\hat{y}^{(i)} \neq y^{(i)}$}:

\State ${c^{n}} = \{ c_{\hat {y}^{(i)}} \mid i=1,\ldots,m \}$

\Else \,\,{$\hat{y}^{(i)} = y^{(i)}$}:
\State compute inter-cluster euclidean distance:  

$\eta_{kj} = \{{\| \sigma(({e^{(i)}_{j}}) - \sigma((c_{c_{k}{j}}) \| _{2}^{2}} \mid i = 1,...,m \,$

\hspace{7mm} $ ; j = 1 , \ldots , C_d$

\hspace{7mm} $; \, \forall \, c_k \mid k \neq y^{(i)}) \, , \, k = 1 , 2 , \ldots , K \}$.
\State  $c^{n}_j = \{ \underset{k \in \{1,2,\ldots,C_d\}}{\text{minimize}} \,\, (\eta_{kj}) \mid j = 1,2,\ldots,C_d\}$
\EndIf
\State compute the triplet center loss by \eqref{AMTCL}.
\EndWhile
\end{algorithmic} 
\end{varalgorithm}\setlength{\textfloatsep}{.2\baselineskip}
In cases where the embedding falls into the region close enough to their class prototype, and far well from the most negative prototype, $\mathcal{L}_{TL} = 0$. Consequently, the sample contribution in backpropagation would be nothing.  During the training process, the AMG  learns to assign an appropriate $\alpha_i$ to samples in a way that it dismissed the samples with simple features and performs hard-mining on samples with the most difficult discriminative features.

To boost the within-cluster compactness, in \eqref{AMTCL} similar to \eqref{vanilla_TL}, we use the cluster centers as the most positive prototype while we synthesize the most negative embedding according to Algorithms~\ref{negative_sample_selection_first}, ~\ref{negative_sample_selection_second}, and ~\ref{negative_sample_selection_third} that correspondingly named mathematically synthesized, network's semantic, and mixed method negative sample selection algorithms.\\
The first algorithm synthesizes the most negative prototype totally based on the euclidean distance, the second one chooses the most negative one for each embedding based on the model output. In misclassified samples, the most negative class prototype
would be exactly the class center of the incorrect prediction. However, for correct predictions, the algorithm looks at the records and selects the class center of the most challenging category that has defeated the model more. The third algorithm is a combination of the two first algorithms, it acts similar to algorithm~\ref{negative_sample_selection_second} for incorrect predictions and algorithm~\ref{negative_sample_selection_first} for correct predictions. 


To elaborate, algorithms~\ref{negative_sample_selection_first} gets all mini batch's embeddings, their corresponding labels, and class center vectors as its inputs and returns a synthesized class center vector as the most negative sample's prototype ${c^{n}_{y^{(i)}}}_{j}$. For the $j$th dimension of the $i$th embedding, the algorithm calculates the Euclidean distance to the $j'$th dimension of all class centers. Then, it selects the smallest one. In this way, the synthesized most negative prototype on each dimension would be a linear combination of all class center vectors projection on that dimension. Note that the coefficients of the linear combination are all zero, and only there is a single one.

Algorithm~\ref{negative_sample_selection_second} acts based on the network predictions. It gets embeddings, their corresponding labels, and the class center matrix as inputs. Then, it compares the predictions and labels and finds the misclassified samples to create a new square matrix of misclassification statistic, $\textbf{S}$, that demonstrates the distribution of misclassified prediction classes for each class label. For incorrectly classified samples, the prediction class center would be considered as the negative sample. However, for the correctly classified ones, the most misclassification of each class during training with that minibatch determines the negative class.

Algorithms~\ref{negative_sample_selection_third}, works similarly to Algorithms~\ref{negative_sample_selection_first} for samples classified correctly, and the same as Algorithms~\ref{negative_sample_selection_second} for samples that are misclassified.




\citet{farzaneh2021facial} argues that not all elements of the embedding have the same effect on the discrimination ability. Inspired by this fact, we believe not all samples' embeddings have the same effect on the discrimination ability. As a result, the AM module provides appropriate $\alpha_i$ to adjust the value of error for each sample as follows:
\begin{align}
\label{AMTCL}
\mathcal{L}_{AMTC3L} = & \frac{1}{2m} \sum_{i=0}^{m} \Big[ \sum_{j=1}^{c_d}  \, \nonumber w_j \Big({\| \sigma({e^{(i)}_{j}}) - \sigma(c^{p}_{y^{(i)}_{j}}) \| _{2}^{2}} \Big) \nonumber\\ 
& -\, w_j \Big({{\| \sigma({e^{(i)}_{j}}) - \sigma(c^{n}_{y^{(i)}_{j}}) \| _{2}^{2}}} \Big) \Big] + \alpha_i,\\
\text{s.t.,}
& \hspace{0.2cm} 0 < \alpha_{i} \leq c_d, 
\label{opt1}
\end{align}
where $\sigma$ represents the Sigmoid function. In this way, the projection of the distance on every single dimension gets a value between zero and one \cite{kertesz2021different}. Hence, if
$\alpha_i = \{\sum_{j=1}^{C_d} w_j \mid w_j \in \{0,1\}\}$ then it assures that in all circumstances $\mathcal{L}_{AMTC3L} > 0$ and the sample would contribute on the backpropagation. However, the number of outlier features that get multiplied by zero cause the margin varies sample by sample.
\vspace{-1 mm}

\begin{table}[!bp]
\centering
\caption{ Performance comparison of AMTC3L-aided CE with ordinary CE and state of the art methods on RAF-DB dataset. Note that our method's results is based on five-fold. cross-validation.}\vspace{0pt}
\label{tab:AchievementComparison}
\renewcommand{\arraystretch}{.005}
\begin{tabular}
{>{\arraybackslash}P{0.75cm}
 >{\arraybackslash}p{1.05cm}
 >{\arraybackslash}p{0.25cm}
 >{\arraybackslash}p{0.25cm}
 >{\arraybackslash}p{0.25cm}
 >{\arraybackslash}p{0.25cm}
 >{\arraybackslash}p{0.25cm}
 >{\arraybackslash}p{0.25cm}
 >{\arraybackslash}p{0.25cm}
 >{\arraybackslash}p{0.5cm} 
  }
\toprule
{\fontsize{6}{5}\selectfont { \rule{-5pt}{0ex}
\textbf{Datasets}
}}
&
{\fontsize{6}{5}\selectfont { \rule{-5pt}{0ex}
\textbf{Models}
}}
&
{\fontsize{6}{5}\selectfont { \rule{-8pt}{0ex}
\textbf{Sur.}
}}
&
{\fontsize{6}{5}\selectfont { \rule{-8pt}{0ex}
\textbf{Fea.}
}}
&

{\fontsize{6}{5}\selectfont { \rule{-8pt}{0ex}
\textbf{Dis.}
}}
&

{\fontsize{6}{5}\selectfont { \rule{-8pt}{0ex}
\textbf{Hap.}
}}
&

{\fontsize{6}{5}\selectfont {\rule{-8pt}{0ex}
\textbf{Sad.}
}}
&

{\fontsize{6}{5}\selectfont {\rule{-8pt}{0ex}
\textbf{Ang.}
}}
&

{\fontsize{6}{5}\selectfont {\rule{-8pt}{0ex}
\textbf{Neu.}
}}
&

{\fontsize{6}{5}\selectfont {\rule{-12pt}{0ex}
\textbf{Average}
}}

\\[-1.5ex]
 \midrule


\multirow{5}{*}{
{\fontsize{6}{6}\selectfont {\rule{-3pt}{0ex}
\textbf{RAF-DB}
}} 
}
&
{\fontsize{6}{6}\selectfont { \rule{-3pt}{0ex}
Baseline
}}
&
{\fontsize{6}{6}\selectfont {\rule{-5pt}{0ex}
0.85
}}
&
{\fontsize{6}{6}\selectfont {\rule{-5pt}{0ex}
0.60
}}
&
{\fontsize{6}{6}\selectfont {\rule{-5pt}{0ex}
0.57
}}
 &
{\fontsize{6}{6}\selectfont {\rule{-5pt}{0ex}
0.94
}}
 &
{\fontsize{6}{6}\selectfont {\rule{-5pt}{0ex}
0.82
}}
 &
{\fontsize{6}{6}\selectfont {\rule{-5pt}{0ex}
0.74
}}
 &
{\fontsize{6}{6}\selectfont {\rule{-5pt}{0ex}
0.86
}}
 &

{\fontsize{6}{6}\selectfont {\rule{-7pt}{0ex}
0.7685
}}

\\[-1ex]
&
{\fontsize{6}{6}\selectfont {\rule{-3pt}{0ex}
MS-NSS
}}
&
{\fontsize{6}{6}\selectfont {\rule{-5pt}{0ex}
0.85
}}
&
{\fontsize{6}{6}\selectfont {\rule{-5pt}{0ex}
0.64
}}
 &

{\fontsize{6}{6}\selectfont {\rule{-5pt}{0ex}
0.59
}}
 &
{\fontsize{6}{6}\selectfont {\rule{-5pt}{0ex}
0.95
}}
 &
{\fontsize{6}{6}\selectfont {\rule{-5pt}{0ex}
0.87
}}
 &
{\fontsize{6}{6}\selectfont {\rule{-5pt}{0ex}
0.81
}}
 &

{\fontsize{6}{6}\selectfont {\rule{-5pt}{0ex}
0.84
}}
&
{\fontsize{6}{6}\selectfont {\rule{-7pt}{0ex}
0.7929
}}
 
\\[-1ex]
&
{\fontsize{6}{6}\selectfont {\rule{-3pt}{0ex}
NS-NSS
}}
&
{\fontsize{6}{6}\selectfont {\rule{-5pt}{0ex}
0.86
}}
&
{\fontsize{6}{6}\selectfont {\rule{-5pt}{0ex}
0.66
}}
 &

{\fontsize{6}{6}\selectfont {\rule{-5pt}{0ex}
0.62
}}
 &
{\fontsize{6}{6}\selectfont {\rule{-5pt}{0ex}
\textbf{0.96}
}}
 &
{\fontsize{6}{6}\selectfont {\rule{-5pt}{0ex}
0.82
}}
 &
{\fontsize{6}{6}\selectfont {\rule{-5pt}{0ex}
0.78
}}
 &

{\fontsize{6}{6}\selectfont {\rule{-5pt}{0ex}
0.86
}}
&
{\fontsize{6}{6}\selectfont {\rule{-7pt}{0ex}
0.7943
}}
 
\\[-1ex]
&
{\fontsize{6}{6}\selectfont {\rule{-3pt}{0ex}
MM-NSS
}}
&
{\fontsize{6}{6}\selectfont {\rule{-5pt}{0ex}
\textbf{0.87}
}}
&
{\fontsize{6}{6}\selectfont {\rule{-5pt}{0ex}
\textbf{0.66}
}}
 &

{\fontsize{6}{6}\selectfont {\rule{-5pt}{0ex}
\textbf{0.64}
}}
 &
{\fontsize{6}{6}\selectfont {\rule{-5pt}{0ex}
0.95
}}
 &
{\fontsize{6}{6}\selectfont {\rule{-5pt}{0ex}
\textbf{0.89}
}}
 &
{\fontsize{6}{6}\selectfont {\rule{-5pt}{0ex}
\textbf{0.80}
}}
 &

{\fontsize{6}{6}\selectfont {\rule{-5pt}{0ex}
\textbf{0.87}
}}
&
{\fontsize{6}{6}\selectfont {\rule{-7pt}{0ex}
\textbf{0.8114}
}}
 
\\[-1ex]

\\[-1ex]
\bottomrule
\end{tabular}
\end{table}
\vspace{0pt}

\subsection{Classifier}
To have the advantage of being FCN, using the average pooling layer in the classifier network is inevitable, $\mathcal{N}_C$. As mentioned, the basic architecture of $\mathcal{N}_C$ consists of two depth-wise $1 \times 1$ kernel-size convolution layers that consecutively reduce the channel size and map the feature space representation, $\mathbb{R}^{C_{f} \times H_{f} \times W_{f}}$, to context space, $\mathbb{R}^{C_{d} \times H_{f} \times W_{f}}$. Then, a global average pooling layer produces embedding tensor, $\mathbb{R}^{C_{d} \times 1 \times 1}$, from context space representation of the input sample. Finally, a fully connected layer maps the embedding to the row score vector.

\begin{table*}[hbt!]
\centering
\caption{ Ablation study}\vspace{5pt}
\label{tab:ablation}
\renewcommand{\arraystretch}{.005}
\begin{tabular}
{ >{\arraybackslash}P{2.8cm}|
 >{\arraybackslash}P{2.8cm}|
 >{\arraybackslash}P{2.8cm}|
 >{\arraybackslash}P{2.8cm}
  }
\toprule
 \centering\fontsize{8}{9}\selectfont\textbf{Settings } & 
 {\fontsize{7}{8}\selectfont {\rule{-5pt}{-0ex}
     \addtolength{\tabcolsep}{-2pt} 
 {\begin{tabular}{@{} cl*{3}c @{}}
  \multicolumn{3}{c}{\centering\fontsize{8}{9}\selectfont\textbf{MS-NSS}} \\[2ex]
  \rot{\centering\fontsize{6}{6}\selectfont\textbf{{Classifier}}} & 
  \rot{\centering\fontsize{6}{6}\selectfont\textbf{{TC3L}}} & \rot{\centering\fontsize{6}{6}\selectfont\textbf{AMTC3L}} 

 \end{tabular}}
 \addtolength{\tabcolsep}{0pt}
 }}&
 {\fontsize{7}{8}\selectfont {\rule{-5pt}{-0ex}
     \addtolength{\tabcolsep}{-2pt} 
 {\begin{tabular}{@{} cl*{3}c @{}}
  \multicolumn{3}{c}{\centering\fontsize{8}{9}\selectfont\textbf{NS-NSS}} \\[2ex]
  \rot{\centering\fontsize{6}{6}\selectfont\textbf{{Classifier}}} & 
  \rot{\centering\fontsize{6}{6}\selectfont\textbf{{TC3L}}} & \rot{\centering\fontsize{6}{6}\selectfont\textbf{AMTC3L}}

 \end{tabular}}
 \addtolength{\tabcolsep}{0pt}
 }}&
 {\fontsize{7}{8}\selectfont {\rule{-5pt}{-0ex}
     \addtolength{\tabcolsep}{-2pt} 
 {\begin{tabular}{@{} cl*{3}c @{}}
  \multicolumn{3}{c}{\centering\fontsize{8}{9}\selectfont\textbf{MM-NSS}} \\[2ex]
  \rot{\centering\fontsize{6}{6}\selectfont\textbf{{Classifier}}} & 
  \rot{\centering\fontsize{6}{6}\selectfont\textbf{{TC3L}}} & \rot{\centering\fontsize{6}{6}\selectfont\textbf{AMTC3L}} 

 \end{tabular}}
 \addtolength{\tabcolsep}{0pt}
  }}

 

\\[4ex]
 \midrule
{\fontsize{7}{6}\selectfont {
baseline
}} &
{\fontsize{7}{8}\selectfont {
\rule{0pt}{-0ex}
    \addtolength{\tabcolsep}{4pt}
  {\begin{tabular}{@{} cl*{3}c @{}}
      {\centering\fontsize{7}{5}\selectfont\textbf{\OK}} &
      {\centering\fontsize{7}{5}\selectfont\textbf{\notOK}} & 
      {\centering\fontsize{7}{5}\selectfont\textbf{\notOK}}  

 \end{tabular}} 
 \addtolength{\tabcolsep}{0pt}
 }} &
{\fontsize{7}{8}\selectfont {
\rule{0pt}{-0ex}
    \addtolength{\tabcolsep}{4pt}
  {\begin{tabular}{@{} cl*{3}c @{}}
      {\centering\fontsize{7}{5}\selectfont\textbf{\OK}} &
      {\centering\fontsize{7}{5}\selectfont\textbf{\notOK}} & 
      {\centering\fontsize{7}{5}\selectfont\textbf{\notOK}}  

 \end{tabular}} 
 \addtolength{\tabcolsep}{0pt}
 }} &
{\fontsize{7}{8}\selectfont {
\rule{0pt}{-0ex}
    \addtolength{\tabcolsep}{4pt}
  {\begin{tabular}{@{} cl*{3}c @{}}
      {\centering\fontsize{7}{5}\selectfont\textbf{\OK}} &
      {\centering\fontsize{7}{5}\selectfont\textbf{\notOK}} & 
      {\centering\fontsize{7}{5}\selectfont\textbf{\notOK}}  

 \end{tabular}} 
 \addtolength{\tabcolsep}{0pt}
 }} 
 

\\[-2ex]
{\fontsize{7}{6}\selectfont {
pipeline a (ablation)
}} &
{\fontsize{7}{8}\selectfont {
\rule{-0pt}{-0ex}
    \addtolength{\tabcolsep}{4pt}
  {\begin{tabular}{@{} cl*{3}c @{}}
      {\centering\fontsize{7}{5}\selectfont\textbf{\OK}} &
      {\centering\fontsize{7}{5}\selectfont\textbf{\OK}} & 
      {\centering\fontsize{7}{5}\selectfont\textbf{\notOK}}  

 \end{tabular}} 
 \addtolength{\tabcolsep}{0pt}
 }} &
{\fontsize{7}{8}\selectfont {
\rule{-0pt}{-0ex}
    \addtolength{\tabcolsep}{4pt}
  {\begin{tabular}{@{} cl*{3}c @{}}
      {\centering\fontsize{7}{5}\selectfont\textbf{\OK}} &
      {\centering\fontsize{7}{5}\selectfont\textbf{\OK}} & 
      {\centering\fontsize{7}{5}\selectfont\textbf{\notOK}}  

 \end{tabular}} 
 \addtolength{\tabcolsep}{0pt}
 }} &
{\fontsize{7}{8}\selectfont {
\rule{0pt}{-0ex}
    \addtolength{\tabcolsep}{4pt}
  {\begin{tabular}{@{} cl*{3}c @{}}
      {\centering\fontsize{7}{5}\selectfont\textbf{\OK}} &
      {\centering\fontsize{7}{5}\selectfont\textbf{\OK}} & 
      {\centering\fontsize{7}{5}\selectfont\textbf{\notOK}}  

 \end{tabular}} 
 \addtolength{\tabcolsep}{0pt}
 }} 

 

\\[-2ex]
{\fontsize{7}{6}\selectfont {
pipeline b (ablation)
}} &
{\fontsize{7}{8}\selectfont {
\rule{0pt}{-0ex}
    \addtolength{\tabcolsep}{4pt}
  {\begin{tabular}{@{} cl*{3}c @{}}
      {\centering\fontsize{7}{5}\selectfont\textbf{\OK}} &
      {\centering\fontsize{7}{5}\selectfont\textbf{\notOK}} & 
      {\centering\fontsize{7}{5}\selectfont\textbf{\OK}}  

 \end{tabular}} 
 \addtolength{\tabcolsep}{0pt}
 }} &
{\fontsize{7}{8}\selectfont {
\rule{0pt}{-0ex}
    \addtolength{\tabcolsep}{4pt}
  {\begin{tabular}{@{} cl*{3}c @{}}
      {\centering\fontsize{7}{5}\selectfont\textbf{\OK}} &
      {\centering\fontsize{7}{5}\selectfont\textbf{\notOK}} & 
      {\centering\fontsize{7}{5}\selectfont\textbf{\OK}}  

 \end{tabular}} 
 \addtolength{\tabcolsep}{0pt}
 }} &
{\fontsize{7}{8}\selectfont {
\rule{0pt}{-0ex}
    \addtolength{\tabcolsep}{4pt}
  {\begin{tabular}{@{} cl*{3}c @{}}
      {\centering\fontsize{7}{5}\selectfont\textbf{\OK}} &
      {\centering\fontsize{7}{5}\selectfont\textbf{\notOK}} & 
      {\centering\fontsize{7}{5}\selectfont\textbf{\OK}}  

 \end{tabular}} 
 \addtolength{\tabcolsep}{0pt}
 }} 

 

\\[-2ex]
\midrule

{\fontsize{7}{6}\selectfont {
baseline (accuracy)
}} &
{\fontsize{7}{6}\selectfont {
\% 85.78
}} &
{\fontsize{7}{6}\selectfont {
\% 85.78
}} &
{\fontsize{7}{6}\selectfont {
\% 85.78
}} 

 

\\[-2ex]
{\fontsize{7}{6}\selectfont {
pipeline a (accuracy)
}} &
{\fontsize{7}{6}\selectfont {
\% 87.12
}} &
{\fontsize{7}{6}\selectfont {
\% 87.29
}} &
{\fontsize{7}{6}\selectfont {
\% 88.53
}} 

 

\\[-2ex]
{\fontsize{7}{6}\selectfont {
pipeline b (accuracy)
}} &
{\fontsize{7}{6}\selectfont {
\textbf{\% 87.77}
}} &
{\fontsize{7}{6}\selectfont {
\textbf{\% 87.81}
}} &
{\fontsize{7}{6}\selectfont {
\textbf{\% 88.90}
}} 

 

\\[-0ex]
\vspace{-0cm}
\\

\bottomrule\vspace{-0cm}
\end{tabular}
\end{table*}\vspace{-0cm}

\subsubsection{Selective attention module}
In general, not all spatial features have the same contribution to the classifying process. 
As \cite{guo2022attention} indicates, every single feature map from channel-wise stacked does not necessarily demonstrate a unique semantic feature. Therefore, leveraging the advantage of the attention layer makes sense. Inspired by \cite{guo2022attention, lotfi2022semantic, lotfi2022evolutionary, xue2021transfer}, a selective inclusion/exclusion attention mechanism is applied to all projections of the embedding vector, anchor, and negative vectors to prevent outliers' contribution to the DML loss function. As mentioned before, this module also defines adaptive margins. \vspace{-0.1cm}

\begin{figure}
     \centering
     \begin{subfigure}[a]{0.18\textheight}
         \centering
         \includegraphics[width=\textwidth]{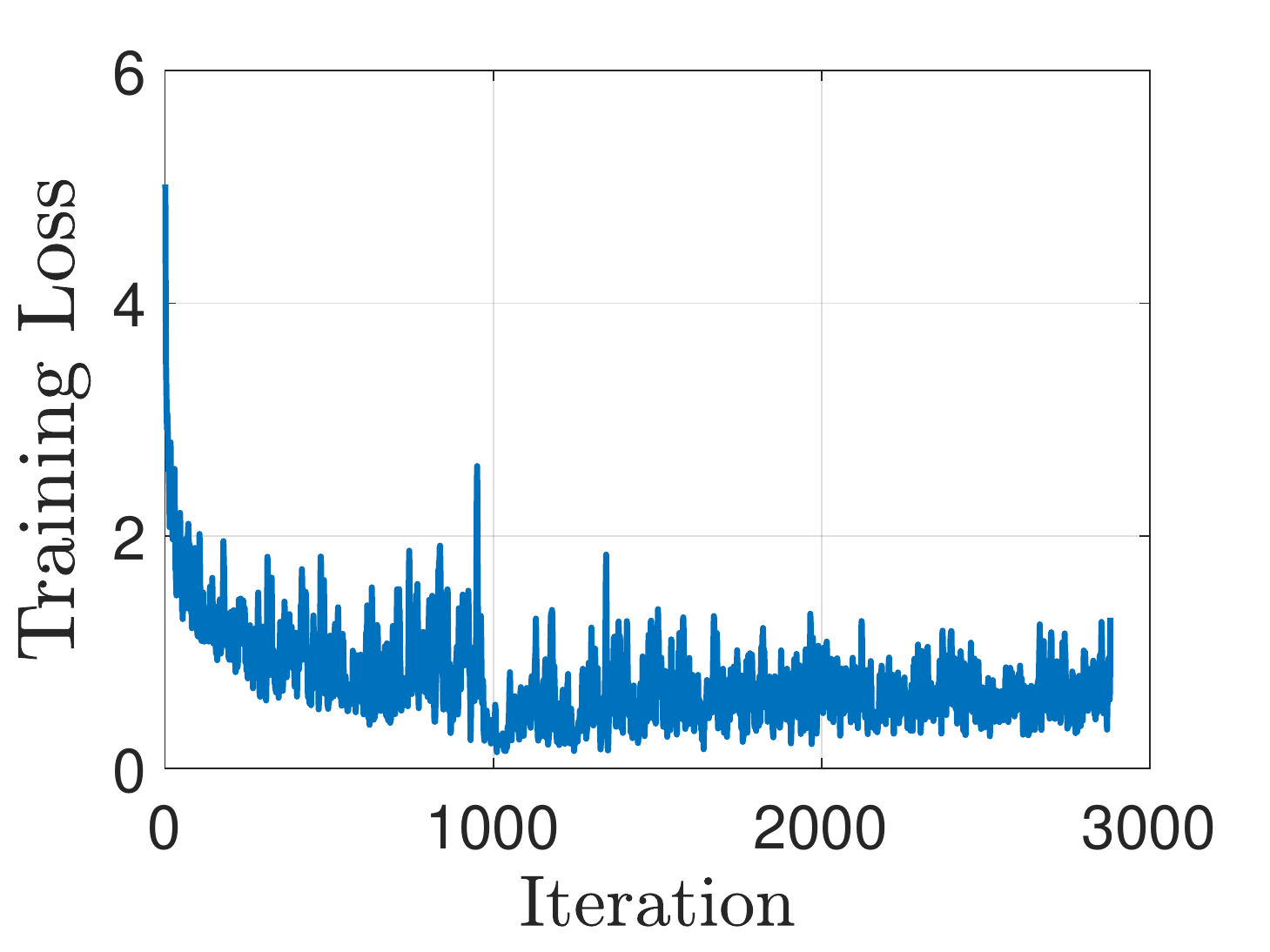}
         \caption{~\ref{negative_sample_selection_first} Convergence}
         \label{convergence_alg1}
     \end{subfigure}
     \hfill
     \begin{subfigure}[a]{0.18\textheight}
         \centering
         \includegraphics[width=\textwidth]{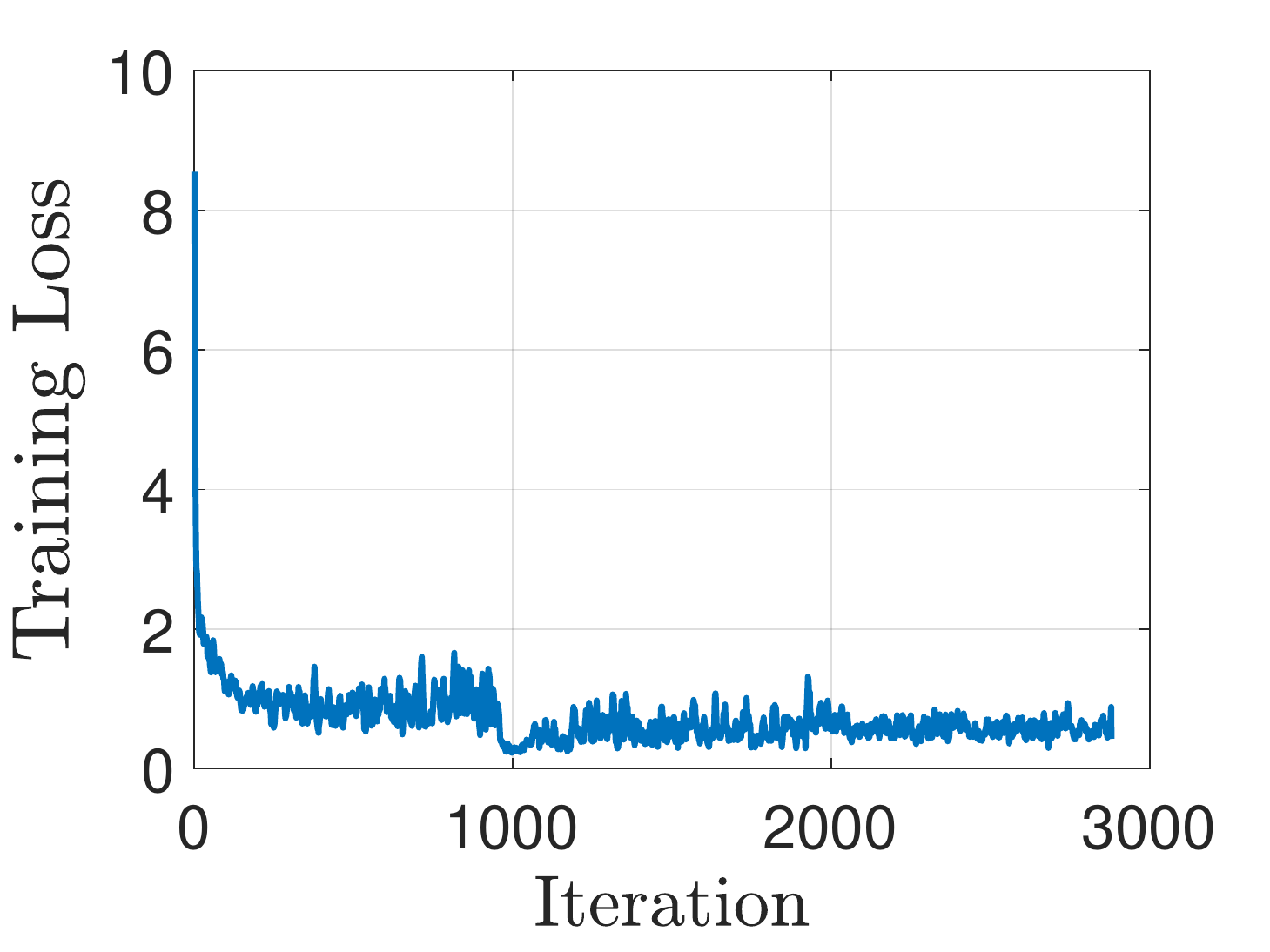}
         \caption{~\ref{negative_sample_selection_second} Convergence}
         \label{convergence_alg2}
     \end{subfigure}
     \hfill
     \begin{subfigure}[a]{0.18\textheight}
         \centering
         \includegraphics[width=\textwidth]{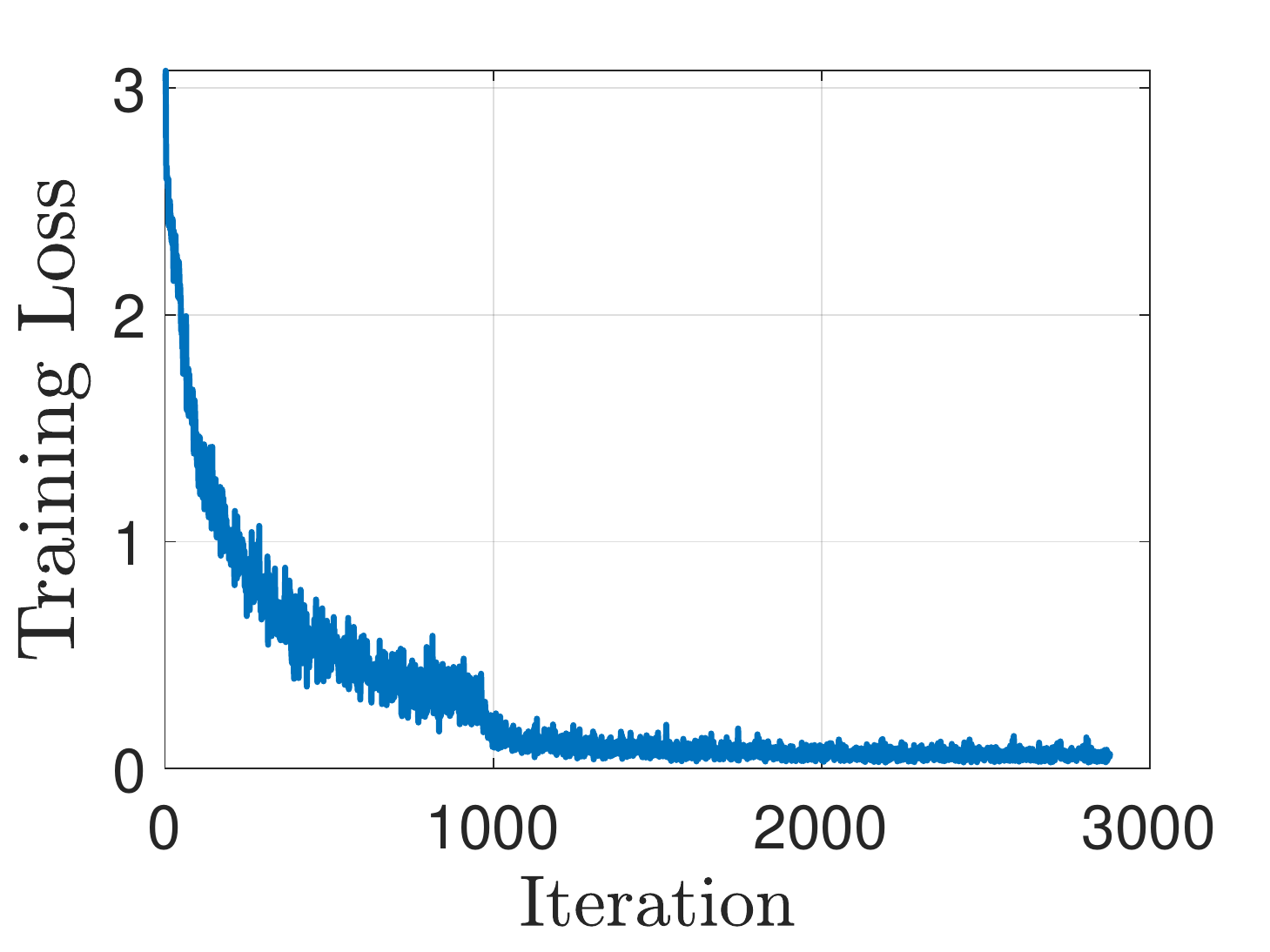}
         \caption{~\ref{negative_sample_selection_third} algorithm Convergence}
         \label{convergence_alg3}
     \end{subfigure}
        \caption{Training loss vs iterations in all three proposed algorithms.}
        \label{conv}
\end{figure}

\section{Experimental Result}\vspace{-0cm}
 In this section, first, we introduce the dataset utilized in this study, the RAF-DB. Then, we compare visually the performance of all three strategies of selecting negative samples that we developed in our work. 
Some samples of RAF-DB has demonstrated in Fig. \ref{fig2}. 


\subsection{RAF-DB}
The Real-world Affective Faces Database (RAF-DB) contains 29672 real-world images with two sets of simple and compound emotions and is considered as one of the richest facial expression databases \cite{li2019reliable}. The first set of labels includes basic emotions like Fearful, Happy, Surprised, Sad, Angry, Disgusted, and Natural. This dataset was gathered in 2017. The dataset's complex labeled emotions included 12 labels such as Fearfully Surprised, Angrily Surprised, and so on. The basic emotions were represented by 15,339 instances of all images, which were divided into 12,271 training samples and 3,068 testing samples. \vspace{-0.1cm}

\subsection{Implementation details}
We used pre-trained Resnet-50 as our backbone, it acts very well in extracting deep features. All proposed algorithms and baseline are trained with standard stochastic gradient descent (SGD) optimizer with the momentum of $0.9$ and the weight decay of $5 \times 10^{-4}$. To augment the input samples we used multi-scale random cropping, random horizontal flipping, random erasing, and normalizing with regard to mean and standard deviation. In order to apply multi-scale random cropping, before feeding a mini-batch of samples to the network we scale them randomly then randomly crop patches with the size of $224 \times 224$ and feed them to the network. In this way, the model can learn local high-discriminative features as well as global features. Note that applying correct augmentation has a significant effect to increase both the baseline and proposed technique. 
 We trained our model on RAF-DB with 60 epochs and a learning rate of 0.01 (only for CE) for baseline and 0.05 (for either CE and AMTC3L) in all three algorithms. The learning rates decay with a factor of 10 every 20 epochs. In our works, the hyper-parameter of $\lambda$ needs to get different values for different algorithms to help the model works at the optimum point. In our architecture, the deep feature is $x^{\ast} \in \mathbb{R}^{2048 \times 4 \times 4}$, and embedding is $e \in \mathbb{R}^{64 \times 1 \times 1}$.

In Fig.\ref{conv} the convergence ability of the proposed algorithms has been demonstrated. Furthermore, in Table \ref{tab:AchievementComparison} and Table \ref{tab:ablation}, the average and overall accuracy have been shown for the proposed algorithms and the baseline. According to Table\ref{tab:ablation}, significant improvement has been achieved by using AMTC3L-aided CE for all proposed algorithms. However, it can get better if we apply the adaptive module scheme that excludes the outliers' effect from backpropagation. As fig.\ref{conv}-\subref{convergence_alg1} demonstrates, there is a huge noisy fluctuation in decreasing loss due to mathematically synthesized negative samples. MS-NSS explores the class centers and builds up single-by-single dimensions of negative samples from the closest elements of other classes. It is the hardest negative sampling scenario among the three proposed approaches. In Algorithm~\ref{negative_sample_selection_second}, smooth convergence comes from the fact that our selection strategy is exactly based on the performance that the model has presented before the current state. There is a tendency to compact/separate intra-/inter- class features but it is not as effective as algorithm~\ref{negative_sample_selection_first}. Algorithm~\ref{negative_sample_selection_first} that takes advantage of both methods shows smooth fluctuation and the fastest convergence.\vspace{-0.1cm}





\section{Conclusion}
In this paper, we have developed a DML attention-aided mechanism that leverages the benefits of the hinge-based loss function, namely TCL. To address the challenge of selecting the most negative samples, we introduced three novel approaches that consider either mathematically the nearest synthesized samples or statistically nearest samples, or a mix of both samples. Accordingly, we realized utilizing the mix of both samples (~\ref{negative_sample_selection_third}) results in a more robust approach. 
The simulation results have shown up to 3.12\% improvement in accuracy compared to the baseline on the RAF-DB dataset. Furthermore, the results have highlighted the importance of a negative sampling strategy in DML-based Loss-less triplet loss scenarios.\vspace{-0.2cm}

{\small
\bibliographystyle{unsrtnat}
\bibliography{egbib}
}

\end{document}